\documentclass{article}

\usepackage{microtype}
\usepackage{graphicx}
\usepackage{booktabs} 

\usepackage{mathtools}
\usepackage{amsmath}
\usepackage{amsfonts}       
\usepackage{multirow}
\usepackage{xspace}
\usepackage{mkolar_definitions}
\usepackage{algorithm}
\usepackage[noend]{algorithmic}
\usepackage{enumitem}
\usepackage{siunitx}
\usepackage{dcolumn}
\usepackage{caption}
\usepackage{subcaption}
\usepackage{float}

\usepackage{hyperref}

\usepackage[accepted]{icml2019}

\newcount\Comments  
\definecolor{purple}{rgb}{0.5,0,1}
\definecolor{teal}{rgb}{0.33,0.65,0.55}
\definecolor{green}{rgb}{0.1,0.65,0.1}
\newcommand{\kibitz}[2]{\ifnum\Comments=1\textcolor{#1}{#2}\fi}

\icmltitlerunning{Supervised Hierarchical Clustering with Exponential Linkage}

\begin{document}

\twocolumn[
\icmltitle{Supervised Hierarchical Clustering with Exponential Linkage}




\begin{icmlauthorlist}
\icmlauthor{Nishant Yadav}{umass}
\icmlauthor{Ari Kobren}{umass}
\icmlauthor{Nicholas Monath}{umass}
\icmlauthor{Andrew McCallum}{umass}
\end{icmlauthorlist}

\icmlaffiliation{umass}{College of Information and Computer Sciences, University of Massachusetts Amherst, USA}

\icmlcorrespondingauthor{Nishant Yadav}{nishantyadav@cs.umass.edu}
\icmlcorrespondingauthor{Ari Kobren}{akobren@cs.umass.edu}
\icmlcorrespondingauthor{Nicholas Monath}{nmonath@cs.umass.edu}
\icmlcorrespondingauthor{Andrew McCallum}{mccallum@cs.umass.edu}

\icmlkeywords{Clustering, Supervised Learning, Machine Learning}

\vskip 0.3in
]



\printAffiliationsAndNotice{}  

\begin{abstract}
  In supervised clustering, standard techniques for learning a pairwise 
  dissimilarity function often suffer from a discrepancy between 
  the training and clustering objectives, leading to poor cluster quality.
  Rectifying this discrepancy necessitates matching the procedure for training the  dissimilarity function to the clustering algorithm. 
  In this paper, we introduce a method for training the 
    dissimilarity function in a way that is tightly coupled with
  hierarchical clustering, in particular single linkage.
  However, the appropriate clustering algorithm for a given 
  dataset is often unknown. Thus we introduce 
  an approach to supervised hierarchical clustering that smoothly 
  interpolates between single, average, and complete linkage, 
  and we give a training procedure that simultaneously learns a linkage function
  and a dissimilarity function.
  We accomplish this with a novel \mbox{Exponential Linkage} function 
  that has a learnable parameter that controls the interpolation.
  In experiments on four datasets, our joint training procedure consistently matches or outperforms the next best training procedure/linkage function pair and gives
  up to 8 points improvement in dendrogram purity over discrepant pairs. 

\end{abstract}

\section{Introduction}
\label{sec:intro}
Clustering algorithms are pervasive in data analysis, pre-processing,
modeling and visualization
~\cite{brown1992class,alon1999broad,seo2002interactively}.  While often
an unsuperivsed problem, in many cases, there exists a small
collection of data points that are labeled with their
\emph{ground-truth} cluster assignments. This enables \emph{supervised
  clustering}, where a dissimilarity function is learned from the
labeled data and then used by a clustering algorithm to partition
unlabeled data~\cite{finley2005supervised}. Supervised clustering is
a common approach in many practical application areas, such as record
linkage~\cite{bilenko2002learning, bilenko2006adaptive}, coreference
resolution~\cite{ng2002improving,
  huang2006efficient,stoyanov2009conundrums} and image segmentation~\cite{liu2013weakly}.

In both supervised and unsupervised settings, the choice of clustering
algorithm bears a significant impact on the quality of the resulting
partition of the data. This is because each clustering algorithm is
designed with specific inductive biases, for example:
$k$-means performs best when
the data points in each cluster are close to that cluster's mean,
DBSCAN~\cite{ester1996density} detects clusters composed of
contiguous, high-density regions, and hierarchical agglomerative
clustering with average linkage discovers clusters for which all
within-cluster data points pairs are similar on average. 
In addition to choosing
a clustering algorithm, supervised clustering demands that
practitioners choose a training procedure for learning the
dissimilarity function that will be used by the algorithm.
Often in practice, the training and clustering objectives are
mismatched, leading to poor cluster quality.

This is especially relevant for the family of hierarchical
agglomerative clustering (\hac) variants, which are widely deployed~\cite{culotta2007author, lee2012joint, levin2012citation,
  kenyon2018resolving}
and are the subject of significant theoretical
study~\cite{dasgupta2016cost, moseley2017approximation}. Variants in the \hac family
iteratively merge
clusters greedily according to the dissimilarity between
pairs of clusters which is measured using a \emph{linkage function}.
\hac variants are differentiated based on the linkage function.
Common linkages, such as
single, average, and complete linkage use pairwise
data point dissimilarities to compute dissimilarities between pairs of clusters. 
In the supervised setting, learning is
typically performed by training the pairwise dissimilarity function to predict
dissimilarity for all within- and across-cluster data points pairs, \emph{regardless} of
the selected linkage function ~\cite{bilenko2002learning, huang2006efficient, stoyanov2009conundrums, guha2015removing}.

However, naively training the dissimilarity function to predict
whether all data point pairs belong to the same cluster does not lead
to robust generalization on unseen data. Especially when clustering is
performed with the single linkage (\slc) variant, which defines the
dissimilarity between two clusters to be their minimum inter-cluster
pairwise dissimilarity, the all-pairs training objective and \slc
constitute a significant mismatch. We demonstrate this phenomenon
empirically and present an algorithm---specifically tailored to \hac
with \slc---for learning dissimilarity functions that empirically
exhibit improved generalization.

Even if equipped with a training algorithm for each \hac variant, it
is often difficult to determine which variant is most appropriate for
a dataset at hand. Ideally, the choice of the variant would be left to a
learning algorithm, which optimizes over a parameterized family of
\hac variants that includes single, average and complete linkage.

In this paper, we establish such a family by expressing each of these
three linkages as a weighted sum of across-cluster
dissimilarities. The weight of an across-cluster dissimilarity is
formed by scaling that dissimilarity by a real-value
hyperparameter, exponentiating the result and normalizing, i.e.,
computing the softmax with respect to the all other pairwise
dissimilarities involved in the merge. We call this family of \hac
linkages the \emph{Exponential linkage (\explink) family}.

We present a training algorithm that jointly selects a linkage from
the \explink family and learns a pairwise dissimilarity function that is suited
to that linkage. The algorithm uses \hac in its inner loop, selecting
specific training examples that promote pure mergers and  
simultaneously optimizes over the \explink family via
gradient descent. 
Crucially, the algorithm obviates the  practitioners' need to 
 choose an appropriate
linkage and a matching training procedure.

We experiment with a cross-product of \hac variants and a variety of
training procedures on four datasets. First, we find that our
specially designed training algorithm for \hac with \slc leads
to improved clustering results over all-pairs training on all
datasets. Furthermore, we find that matching the training algorithm
and \hac variant leads to improved performance over
linkage-agnostic training procedures, such as all-pairs, by up to 8 points of
dendrogram purity. Finally, the results
reveal that our joint training procedure outperforms or matches the
performance of the next best training-procedure/linkage function
combination on four datasets. We highlight this result as being
useful, especially in practice, because it renders selection of the
linkage function before training unnecessary.

\section{Supervised Clustering}
\label{sec:problem}

Clustering is the problem of partitioning a dataset into disjoint
subsets. Let
$\data = \{ x_i \}_{i=1}^m, \data \subset \Xcal$, be a dataset of $m$ points. A \emph{clustering} of $\data$ is a
collection of disjoint subsets (i.e., clusters) of $\data$,
$\mathcal{C} = \{C_i\}_{i=1}^{K'}$, such that
$ \bigcup_{i=1}^{K'} C_i = \data$. The clustering of $\data$ is computed
by a clustering algorithm, $\Acal$. Often the algorithm
makes use of a function,
$f_{\theta}: \Xcal\times\Xcal \rightarrow \RR$, which computes the
dissimilarity of data point pairs in $\Xcal$. We assume that there
exists a \emph{ground-truth} clustering,
$\Ccal^{\star} = \{C^\star_i\}_{i=1}^{K}$ of $\data$.

In supervised clustering, the input dataset has associated labels,
which are used to produce an algorithm to accurately cluster unseen
data.
\begin{definition}\textbf{(Supervised Clustering)}
  Let $\Scal = \{(\data_1,\labels_1) \ldots (\data_n,\labels_n)\} \subset 2^{\Xcal}\times\Ycal$ be a
  training set, where each $\data_i = \{x_j\}_{j=1}^{m_i} \subset \Xcal$ is a
  collection of $m_i$ points and $\labels_i \in \Ycal$ encodes the ground-truth
  partition of $\data_i$. The goal is to learn an algorithm
  $\Acal: 2^{\Xcal} \rightarrow \Ycal$ that accurately clusters a
   new dataset $\data$, where $\forall \ \ 1\leq i \leq n, $ $ \data \cap \data_i = \emptyset$ ~\cite{finley2005supervised}.
\end{definition}
In practice and in previous work, rather than learning $\Acal$
directly, a dissimilarity function $f_\theta$ is learned instead and
used to construct $\Acal$~\cite{finley2005supervised,culotta2007author,lee2012joint,levin2012citation,kim2016random,kenyon2018resolving}.

This work focuses on supervised \emph{hierarchical} clustering,
because of its wide usage in practice. A hierarchical clustering
algorithm is one that returns a tree structure for which 
each leaf 
corresponds to a unique data point and 
each internal node corresponds
to the cluster of its descendant leaves.
 Apart from facilitating data
exploration and analysis \cite{seo2002interactively}, the primary
advantage of hierarchical clustering over \emph{flat} clustering is
that the tree simultaneously represents multiple alternative flat
clusterings of a dataset, known as \emph{tree consistent
  partitions}~\cite{heller2005bayesian}. This alleviates the
requirement of specifying the number of clusters \emph{a priori}. In
supervised hierarchical clustering, data labels correspond to a flat
clustering, as in the non-hierarchical setting, and provide signal to
discover trees that encode high quality tree consistent partitions.

\subsection{Hierarchical Agglomerative Clustering}
Hierarchical agglomerative clustering (\hac) is an iterative algorithm
that builds a tree, $\Tcal$, over a dataset one node at a time,
according to a \emph{linkage function}. A linkage function
$l: 2^{\Xcal} \times 2^{\Xcal} \rightarrow \mathbb{R}$ scores the
merger of two \emph{nodes}, where each node corresponds to a cluster
containing data points stored at its descendant leaves.  The algorithm
is initialized by creating one node for each data point. 
The algorithm proceeds in a series of \emph{rounds}.
In each round
of \hac, the two nodes that minimize the linkage function are
\emph{merged}, by making them siblings of one another and creating a
new node to serve as their parent. The algorithm terminates after the
final merge, which creates the root of the tree. \hac has enjoyed
significant study in the theoretical community and usage by
practitioners~\cite{eisen1998cluster, diez2015novel, yim2015hierarchical,
  gan2015faster, xu2016michac, moseley2017approximation, 
  ieva2018discovering, tie2018application}.

\section{Illustrative Example}
The clustering constructed via \hac depends on the choice of both the linkage
function and the pairwise dissimilarity function that powers the
linkage function. Given the opportunity to learn the dissimilarity function,
the training objective optimized during learning should, in some
sense, ``match'' the chosen linkage. In this section, we empirically
show that mismatch between training and clustering objective can
result in poor clustering performance. Our example centers on \hac
with single linkage (\slc).

\subsection{Single Linkage}
\label{subsec:sl}
\slc computes the dissimilarity between two nodes in $\Tcal$ as the
minimum dissimilarity among their corresponding data points.  Let $v,
v' \in \Tcal$ be nodes in the tree and let $C, C'$ be the sets of data points
corresponding to the descendant leaves of $v$ and $v'$, respectively.
\slc is computed as follows: $
l_{\slc}(v,v';\theta) = \min_{ (x_i,x_j) \in C \times C'}
  f_{\theta}(x_i,x_j)$.
\slc is closely related to $\data$'s minimum spanning tree (MST).
Consider a clique-structured graph with the data points as nodes and
the weight of each edge equal to the dissimilarity of its endpoints, 
the edges in the MST of $\data$ correspond to the sequence of
mergers performed during an invocation of \slc (ignoring
ties)~\cite{gower1969minimum}. \slc clustering does not require that
all pairs of points within a cluster be similar; rather, for each pair of points in a cluster there must exist 
a low dissimilarity path. This characteristic of \slc facilitates discovery of clusters with arbitrary structure, but also
causes its sensitivity to outliers.

\subsection{All-pairs Training}
\label{subsec:ap-for-sl}
A commonly used  objective for training
dissimilarity functions for supervised (hierarchical) clustering is
classification loss on all within- and across-cluster data point
pairs~\cite{bilenko2002learning, huang2006efficient, stoyanov2009conundrums, guha2015removing}, which we will call
\emph{all-pairs} (\ap). Let $\data = \{x_i\}_{i=1}^m$ be a set of
points with ground-truth clusters, 
$\Ccal^\star=\{C^\star_i\}_{i=1}^K$. 
Let $\mathbf{x_{i,j}}=(x_i,x_j)$ be a pair of points and let $\Wcal_{\Ccal^\star}, \Acal_{\Ccal^\star}$ be set of within- and across-cluster pairs of points w.r.t $\Ccal^\star$, respectively . 
Define all pairs (\ap) loss as
\begin{equation}
  J_{\ap}(\theta;\Ccal^\star) =  \sum_{\mathbf{x_{i,j}} \in \Wcal_{\Ccal^\star}}  f_{\theta}(\mathbf{x_{i,j}}) \
   - \sum_{ \mathbf{x_{k,l}} \in \Acal_{\Ccal^\star} } f_{\theta}(\mathbf{x_{k,l}})
\end{equation}
However, this objective has a mismatch with \slc. Under this
training objective, misclassifying within- and across-cluster pairs is
equally penalized. Yet, misclassified across-cluster pairs are
precisely the outliers that lead to the demise of \slc. Moreover, when
clustering with \slc, it is possible to recover the ground-truth
partition of a dataset even if the dissimilarity function outputs high
dissimilarity for some within-cluster pairs.

\begin{figure*}[!ht]
\centering
\begin{subfigure}[t]{.3\textwidth}
	{
		\includegraphics[scale=0.3]{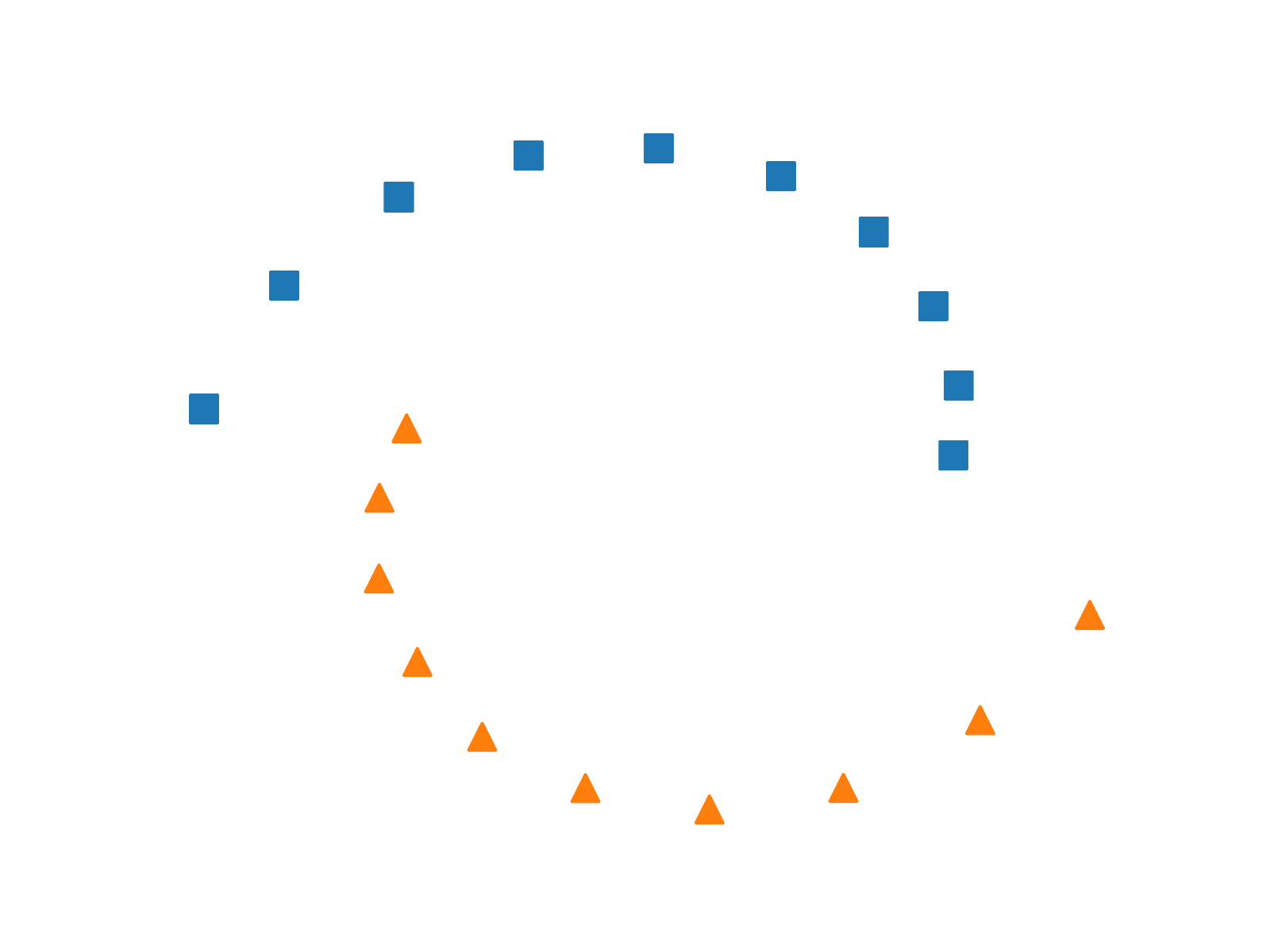}
		\caption{Two clusters in $\RR^2$}
		\label{fig:gtSynth}
	}
\end{subfigure}
\begin{subfigure}[t]{.3\textwidth}
{
	\includegraphics[scale=0.3]{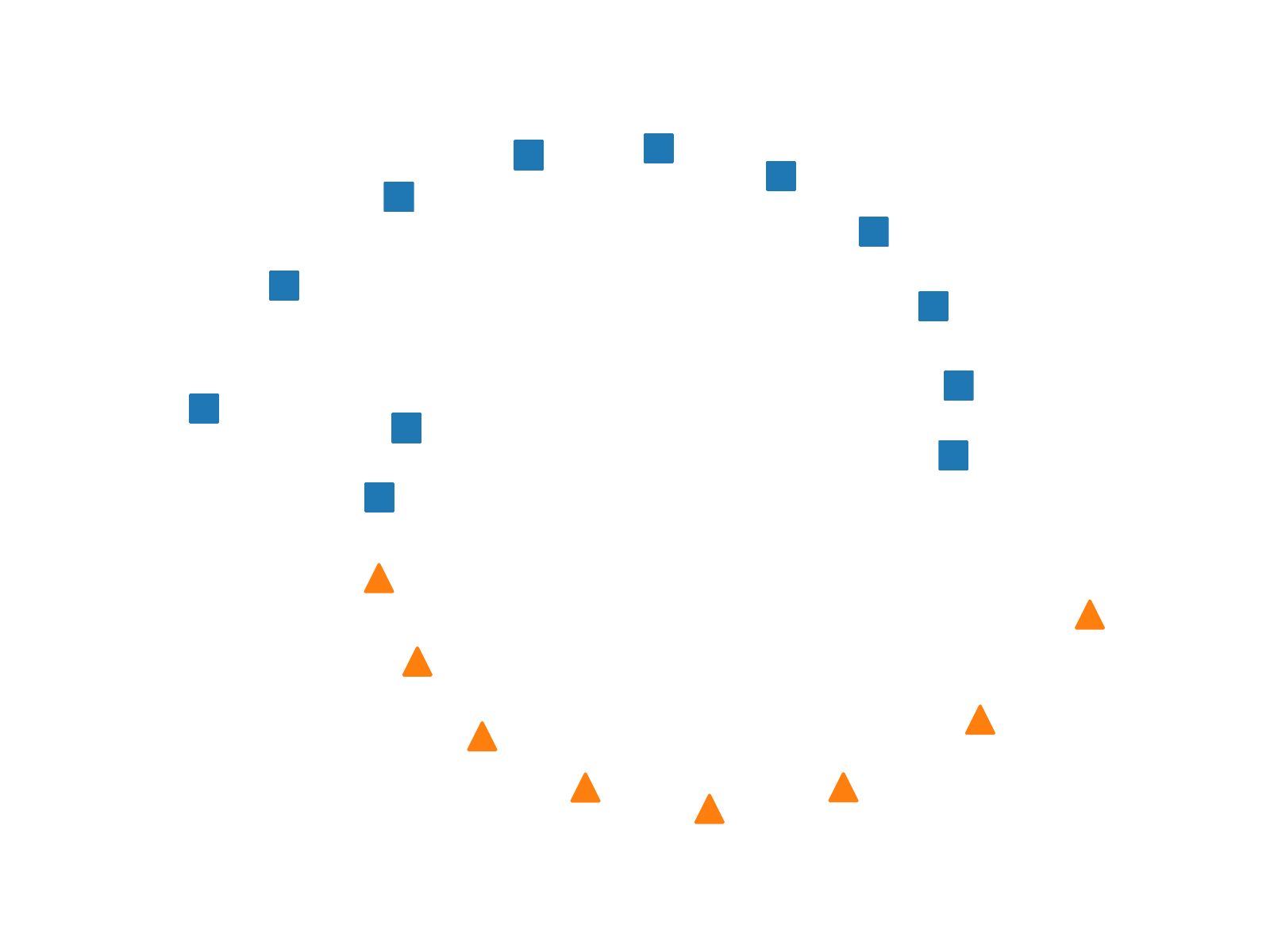}
	\caption{Imperfect clustering}
	\label{fig:allPairSynth}
}
\end{subfigure}
\begin{subfigure}[t]{.3\textwidth}
{
  \includegraphics[scale=0.35]{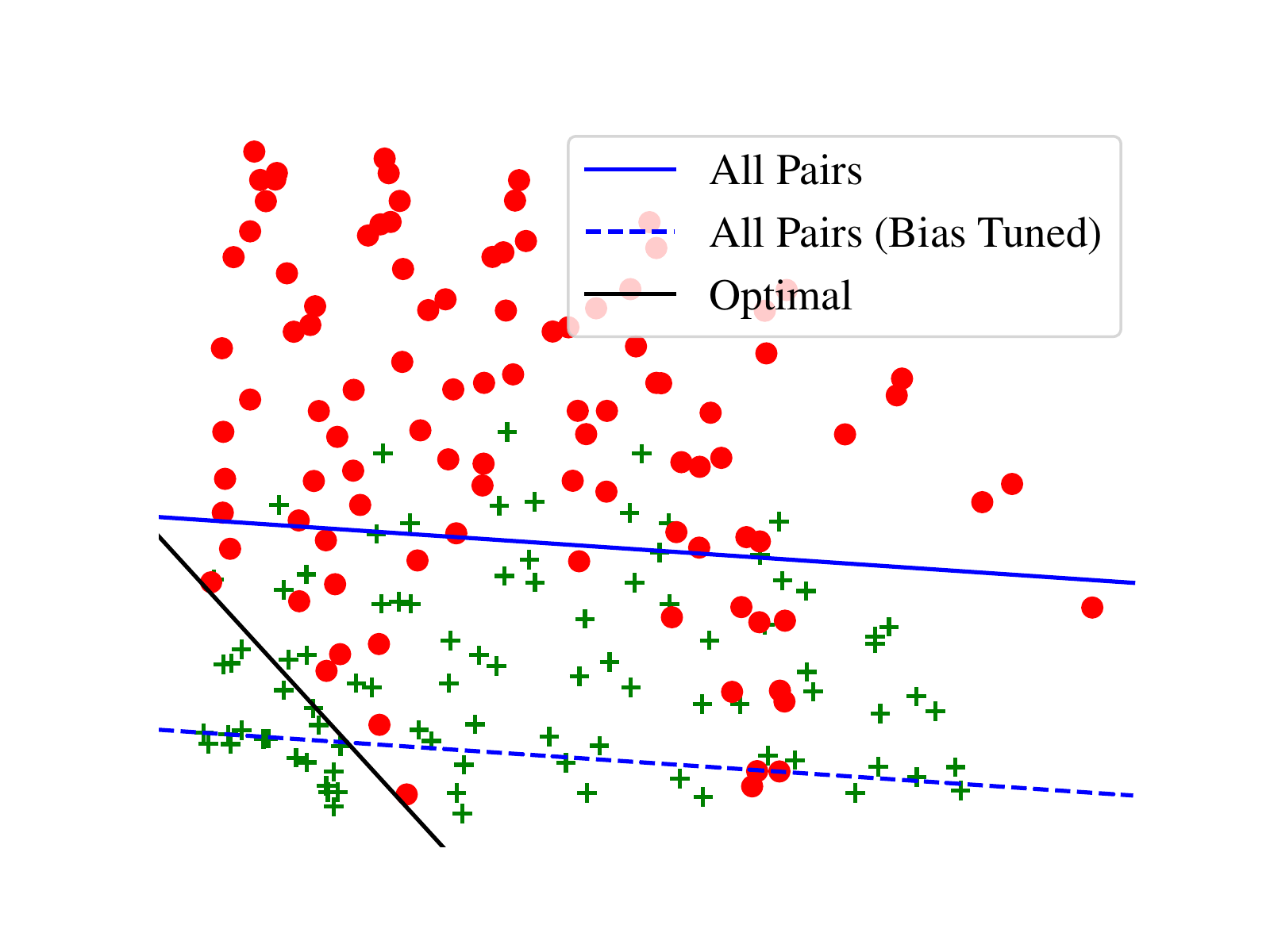}
  \caption{Decison Boundaries}
  \label{fig:OptDecisionSynth}
}
\end{subfigure}
\caption{\textbf{\slc discrepancy with \ap}. Figure~\ref{fig:gtSynth} and~\ref{fig:allPairSynth} depict a
  dataset and the clustering achieved by training the dissimilarity 
  function with \ap followed by clustering with \slc and extraction
  of flat clustering that minimizes errors. Figure~\ref{fig:OptDecisionSynth} shows the absolute difference
  of within-(green) and across-cluster(red) data points pairs along with
  decision boundary learned by \ap, the decision boundary after 
  tuning bias to minimize clustering errors,  
  and the optimal decision boundary. The learned boundary
  minimizes classification error of within- and across-cluster pairs, while the optimal boundary does not misclassify any
  across-cluster pairs but misclassifies many within-cluster pairs.}

\end{figure*}
We present an example showing that \ap training leads to an ineffective dissimilarity function for \slc clustering.
Fig.~\ref{fig:gtSynth} shows a dataset in $\RR^2$ with two
ground-truth clusters, differentiated by color. We use \ap to train a
linear dissimilarity function
$f_{\theta}(x, x') = \theta_{1:2}^{T} \lvert x - x' \rvert + \theta_0$, and
then use $f_\theta$ to cluster via \hac with
\slc.

Figure~\ref{fig:allPairSynth} contains the result, which shows that
the learned function leads to an imperfect clustering.  We plot the
learned decision boundary, the decision boundary
after tuning the bias to minimize clustering 
errors, the optimal decision boundary, and the
vectors $\lvert x - x' \rvert, \forall x,x' \in \data$ 
(Figure \ref{fig:OptDecisionSynth}). The figure shows that
the decision boundary learned by \ap is significantly different from
the optimal decision boundary. Visual inspection anecdotally confirms
that \ap training gives rise to a decision boundary that minimizes
overall classification error while the optimal decision boundary does
not tolerate any misclassifications of across-cluster pairs, while
allowing many within-cluster pairs to be misclassified. Note that it
may be possible to use \ap with a more complex model to avoid some
clustering errors, but employing such a model is more prone to
overfit the training set, which is a primary concern in the
supervised clustering regime.

\subsection{An MST-based Training Algorithm}
\label{subsec:sl-learn}
In training a dissimilarity function for \slc, the goal should be to
minimize misclassifications of across-cluster pairs while retaining
the ability to recognize a sufficient number of within-cluster
pairs. With these considerations in mind, we propose an MST-based loss
function for learning a dissimilarity function that is well-matched
with \slc. At a high-level, minimizing the loss amounts to learning a
low cost MST for each ground-truth cluster while maximizing dissimilarity
of across-cluster edges. Note that, like \slc, the loss only requires
that a handful of within-cluster data point pairs be similar.

Let $\data = \{x_i\}_{i=1}^m, \data \subset \Xcal$, be a
dataset with ground-truth clustering, $\Ccal^{\star} = \{C^\star_i\}_{i=1}^{K}$.
Let $f_{\theta}: \Xcal\times\Xcal \rightarrow \RR$ be a function,
parameterized by $\theta$, that computes the dissimilarity of a pair
of points in $\Xcal$. Let $\texttt{MST}_\theta(C)$ return pairs of points that correspond to
the endpoints of the edges in the MST of ground-truth cluster $C$,
with respect to the dissimilarity function $f_\theta$. Then, define
the following loss:
\vspace{-5mm}

{\footnotesize
\begin{equation}\label{eq:mstLoss}
J_{\slc}(\theta;\Ccal^\star)\! =\!\! \sum_{C \in \Ccal^{\star}}\!\!\Bigg( \!\smashoperator[r]{\sum_{\mathbf{x_{i,j}} \in \texttt{MST}_\theta(C)}} f_{\theta}(\mathbf{x_{i,j}}) \ 
-\! \sum_{x_k \in C} \min_{x_l \not\in C} f_{\theta} (\mathbf{x_{k,l}})\!\Bigg)
\end{equation}
}
This loss can be minimized iteratively. During \mbox{iteration $t$}, construct an
MST for each ground-truth cluster with respect to $f_{\theta^{(t)}}$ and
take gradient steps to minimize the corresponding dissimilarities. For
each ground-truth cluster $C \in \Ccal^\star, \forall x \in C$, find the point
$x' \in \data \setminus C$ that is least dissimilar to $x$, and
take a gradient step to increase the corresponding dissimilarity.
Pseudocode appears in Algorithm~\ref{alg:genTrainExSingle}.
\setlength{\textfloatsep}{2pt}
\begin{algorithm}[t]
   \caption{\texttt{train\_SL}$(\data, \Ccal^\star,T,\gamma)$}
   \label{alg:genTrainExSingle}
\begin{algorithmic}
	\STATE \textbf{Init}: $\theta$
	\FOR{$t = 1,\ \dots,\ T$}
		\STATE $J \gets 0$
		\FOR{$C \in \Ccal^\star$}
			\FOR{$(x_i,x_j) \in \texttt{MST}_\theta(C)$}
				\STATE $J \gets J + f_\theta({x_i,x_j})$
			\ENDFOR
			\FOR{$x \in C$} 
				\STATE $J \gets J - \min_{x' \in \data \setminus C} f_\theta(x,x')$
			\ENDFOR
		\ENDFOR
		\STATE $\theta \gets \theta - \gamma \frac{\partial J}{\partial \theta}$
	\ENDFOR
\end{algorithmic}
\end{algorithm}

%
%
%

A variant of the loss function in Equation$~\ref{eq:mstLoss}$ can be
obtained by using a threshold $\tau \in \RR$, and margin $\mu \in \RR$.
In this case, loss is only incurred when pairs of within-cluster
data points in the MST have dissimilarity greater than $\tau - \mu$ or
when across-cluster pairs have dissimilarity less than $\tau + \mu$.

As anecdotal evidence of this training paradigm, we note that the
optimal decision boundary depicted in
Figure~\ref{fig:OptDecisionSynth} was, in fact, learned using this
training procedure.

\section{Exponential Linkage Clustering}
\label{sec:explink}

While we provided a specialized training
procedure for \slc that leads to better generalization, in general,
the most suitable linkage function for a dataset is, \emph{a priori}, unknown.

In this section, we introduce \Exponential linkage (\explink), a
parametric family of linkage functions for \hac that smoothly
interpolates between single, average and complete linkage--three
widely used linkage functions. We begin by formally defining
\explink~(\S\ref{subsec:expfun}). We present an example
illustrating the advantage of a linkage from \explink over standard linkages~(\S\ref{subsec:explink-eg}). 
Then, we present a training
procedure for jointly learning the interpolation parameter of \explink and the dissimilarity function~(\S\ref{subsec:expfun_train}).

\subsection{Exponential Linkage Function}
\label{subsec:expfun}

Let $C_u,C_v \subset \Xcal$, and let $\mathbf{C}_{u,v} = (C_u,C_v)$.
We define \mbox{\Exponential linkage} (\explink) as:
\begin{equation}
\expLinkFunc{\alpha}(\mathbf{C}_{u,v}) = \frac{\sum\limits_{\mathbf{x_{i,j}}  \in C_{u}\times C_{v}} e^{\alpha f(\mathbf{x_{i,j}}) } f(\mathbf{x_{i,j})}} {\sum\limits_{\mathbf{x_{i,j}}  \in C_{u} \times C_{v}} e^{\alpha f(\mathbf{x_{i,j}}) }}
\end{equation}
where $\alpha \in \RR$ is a hyperparameter that
interpolates between members of the \explink family.

\explink computes the dissimilarity between two clusters via a
weighted average of all inter-cluster pairwise dissimilarities.  Note
that as $\alpha \rightarrow -\infty$, $\expLinkFunc{\alpha}$
approaches the minimum pairwise dissimilarity among its arguments,
i.e., \slc. Similarly, when $\alpha \rightarrow \infty$,
$\expLinkFunc{\alpha}$ approaches complete linkage and when
$\alpha = 0$, $\expLinkFunc{\alpha}$ is average linkage.

Different members of the
\explink family encode different inductive biases with respect to which
pairwise dissimilarities are most important when calculating
dissimilarity between two groups of data points.

\subsection{Illustrative Synthetic Example}
\label{subsec:explink-eg}
\begin{figure*}[ht!]
\centering
\begin{subfigure}[t]{.195\textwidth}
{
	\includegraphics[scale=0.25, trim={2cm 1cm 2cm 1cm},clip]{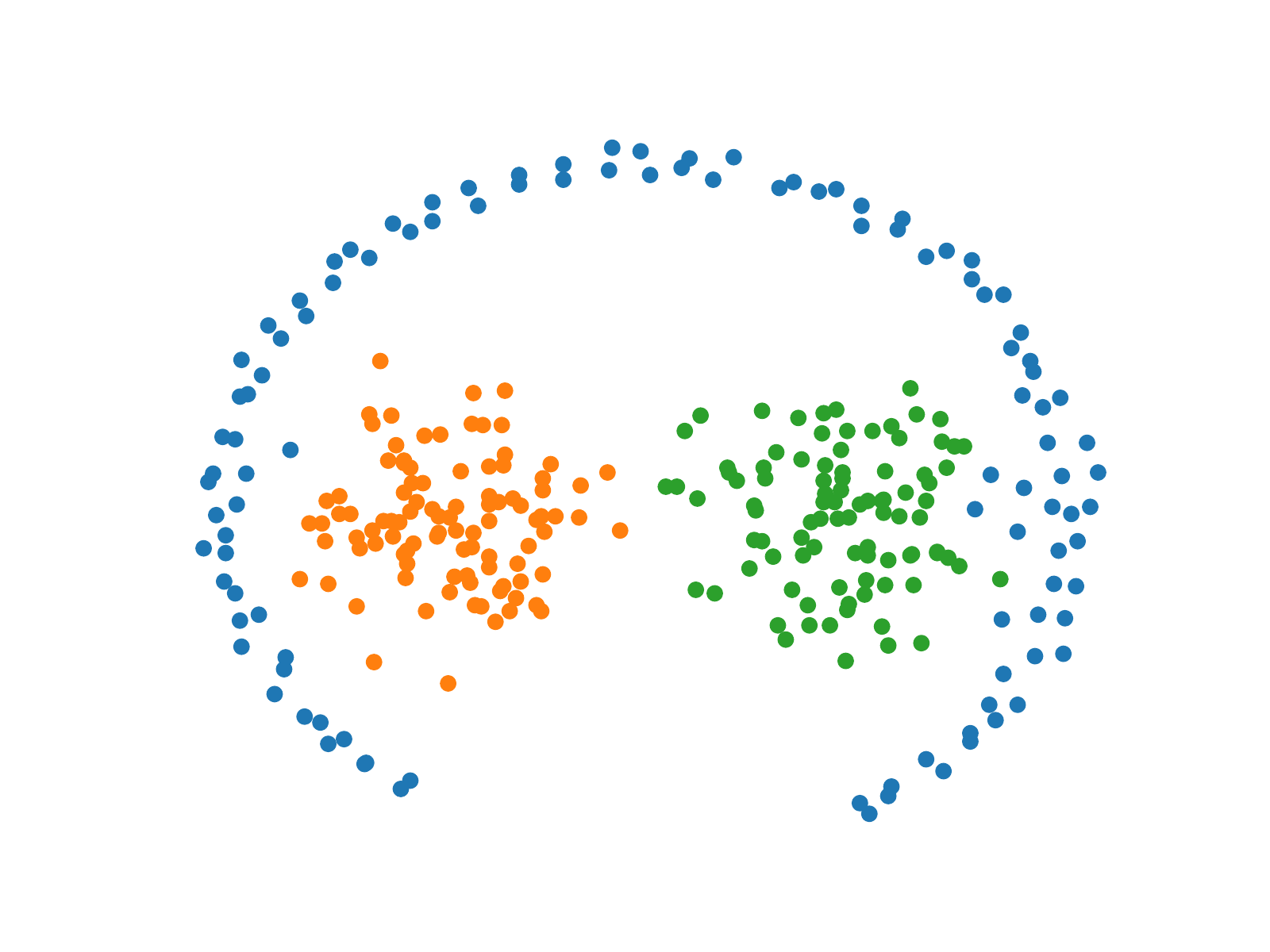}
	\caption{Ground-Truth Clusters}
	\label{fig:pathOriginal}
}
\end{subfigure}
\begin{subfigure}[t]{.195\textwidth}
{
	\includegraphics[scale=0.25, trim={2cm 1cm 2cm 1cm},clip]{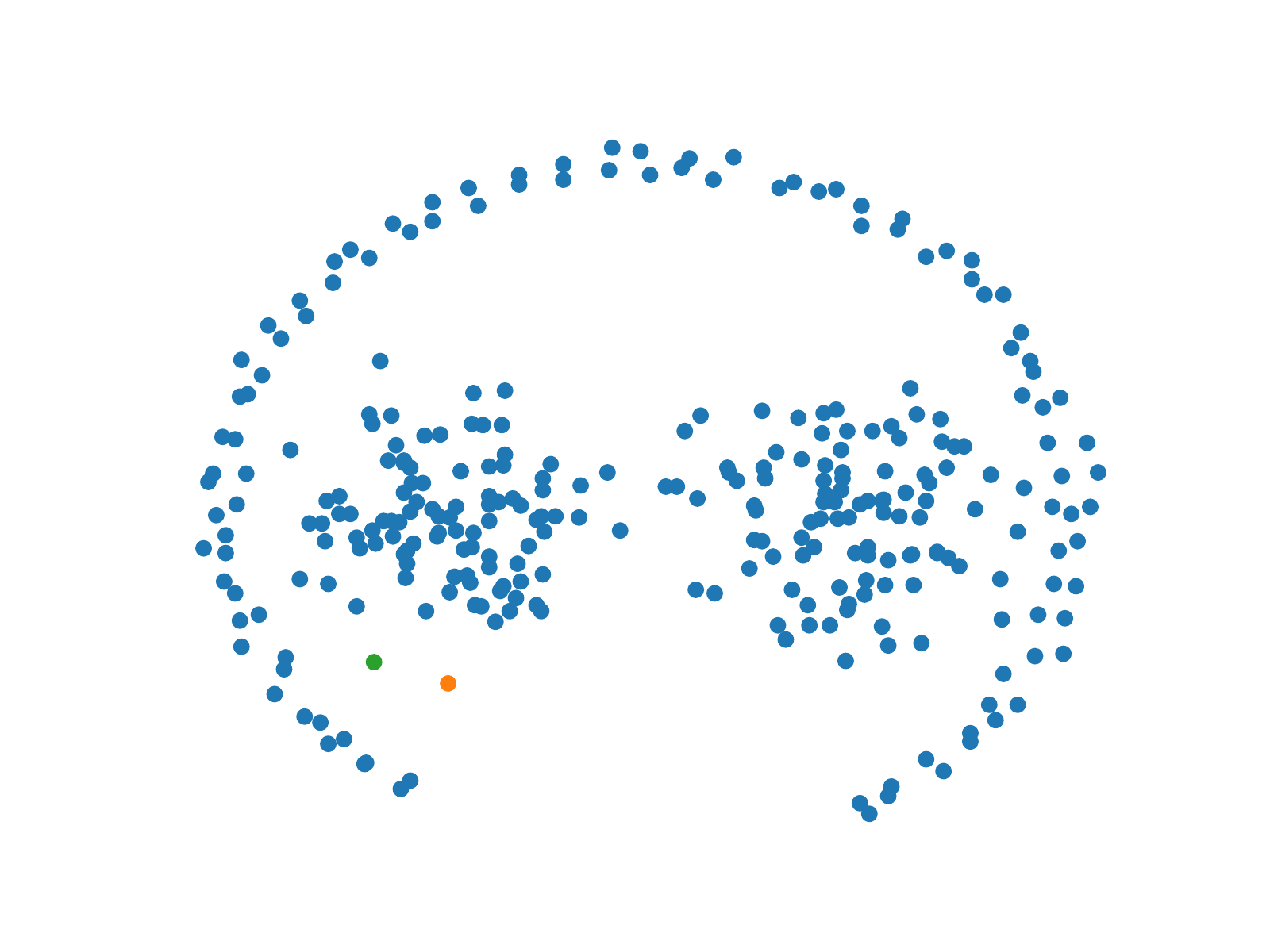}
	\caption{Single Linkage }
	\label{fig:pathSingle}
}
\end{subfigure}
\begin{subfigure}[t]{.195\textwidth}
{
	\includegraphics[scale=0.25, trim={2cm 1cm 2cm 1cm},clip]{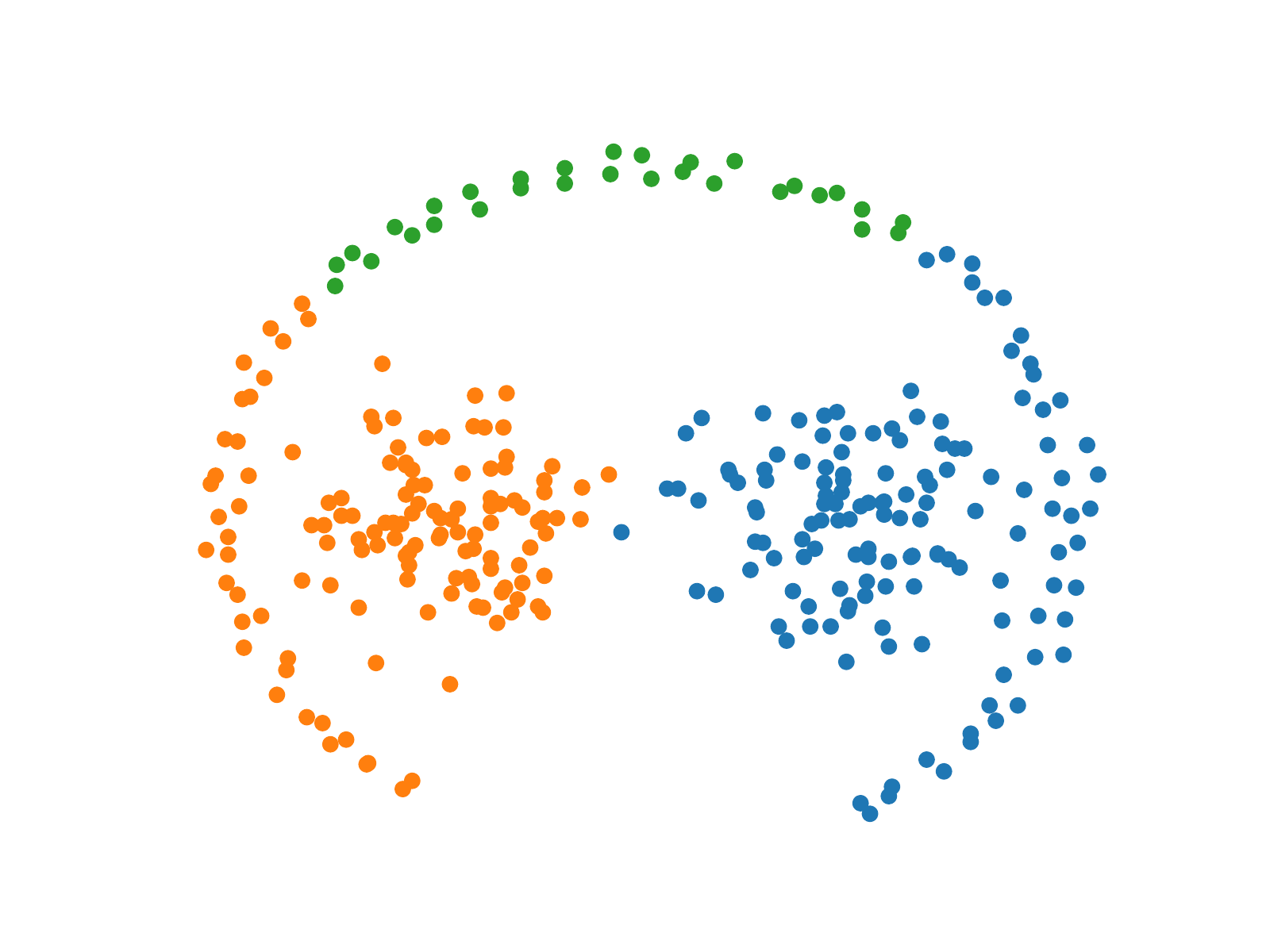}
	\caption{Average Linkage}
	\label{fig:pathAverage}
}
\end{subfigure}
\begin{subfigure}[t]{.195\textwidth}
{
	\includegraphics[scale=0.25, trim={2cm 1cm 2cm 1cm},clip]{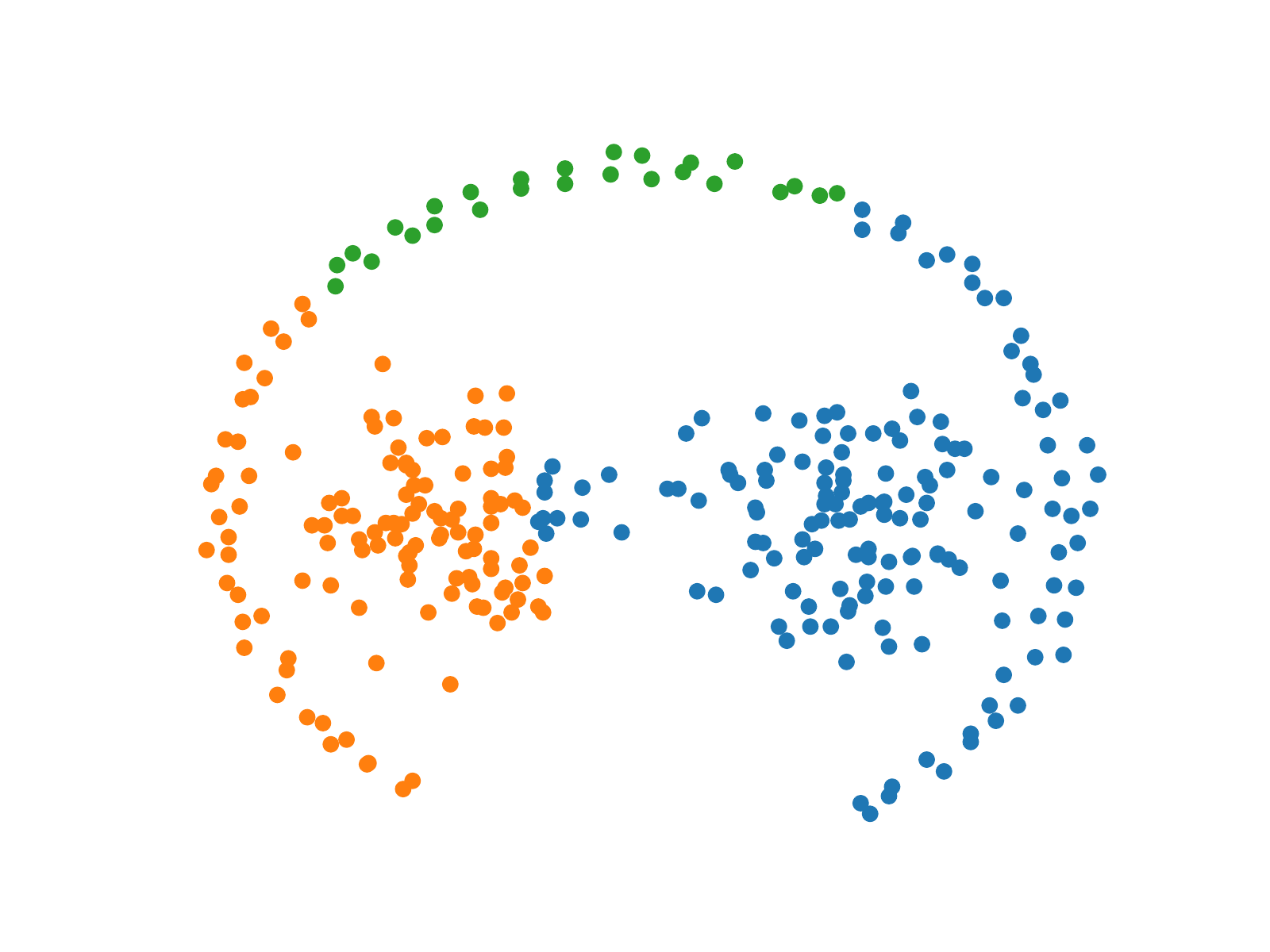}
	\caption{Complete Linkage}
	\label{fig:pathComplete}
}
\end{subfigure}
\begin{subfigure}[t]{.195\textwidth}
{
	\includegraphics[scale=0.25, trim={2cm 1cm 2cm 1cm},clip]{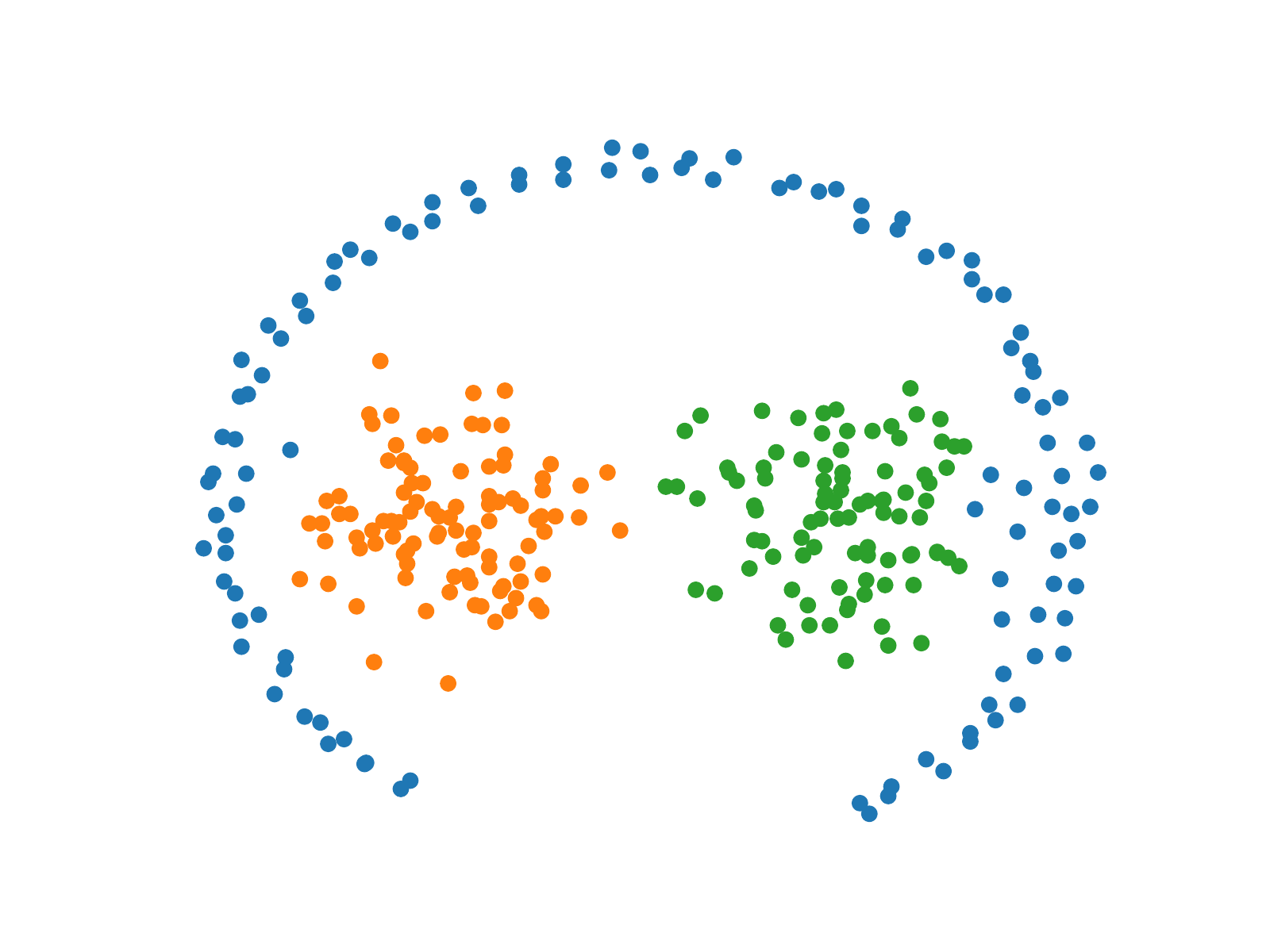}
	\caption{\explink {\small with} $\alpha = -1$ }
	\label{fig:pathExpLink}
}
\end{subfigure}
\caption{Fig.~\ref{fig:pathOriginal} shows three clusters differentiated by color. Fig.~\ref{fig:pathSingle},~\ref{fig:pathAverage},~\ref{fig:pathComplete},~\ref{fig:pathExpLink} shows three flat clusters obtained with \hac using single, average, complete linkage, and \explink with $\alpha=-1$ respectively.  All of the standard linkages fail to recover the ground-truth clusters while \explink with $\alpha=-1$, which corresponds to a linkage function that assigns higher weight to lower dissimilarity edges, is able to recover the desired clustering.}
\label{fig:pathbased}
\end{figure*}

In this section, we demonstrate the flexibility of \explink family.
Figure~\ref{fig:pathbased} shows a synthetic dataset in $\RR^2$
containing three ground-truth clusters. \hac is run with single,
average, complete linkage, and \explink with $ {\footnotesize \alpha\!\!=\!\!-1 }$ 
with the dissimilarity function set to Euclidean distance. 
Flat clusters are obtained by cutting the resultant trees to
obtain three clusters. The dataset is interesting because it contains
both spherical clusters with noisy boundaries and
non-spherical cluster composed of a contiguous, high-density 
region. Each of the standard linkage functions fail to
recover ground-truth clusters for different reasons. \slc
fails because of the chaining effect~\cite{janowitz1978order}
on noisy cluster boundaries. Average and
complete linkages fail because they tend to discover spherical clusters.
However, \explink with $\alpha\!=\!-1$, which assigns higher weight to less 
dissimilar data point pairs, recovers
the ground-truth clusters. This anecdotal evidence suggests that various
members of the \explink family are capable of recovering clusters of complex shapes.

\subsection{Training \explink}
\label{subsec:expfun_train}

As above, let $\data = \{x_i\}_{i=1}^{m}$ be a set of data points
with ground-truth clusters
$\Ccal^\star = \{C^\star_i\}_{i=1}^K$. Let
$\{\Tcal_k^{(i)}\}_{k=0}^{l_i}$ be the set of trees constructed before round
$i$ of \hac. Denote the cluster comprised of the leaves of tree
$\Tcal_k^{(i)}$ by $C_k^{(i)}$, and let $\Ccal^{(i)}$ = $\{C_k^{(i)}\}_{k=0}^{l_i}$. Finally, let $\Pcal^{(i)}$ be
the set of cluster pairs before round $i$, $\Pcal_{+}^{(i)}$
be the set of pair of clusters in $\Pcal^{(i)}$ which are subsets of the same ground-truth
cluster, and $\Pcal_{-}^{(i)}$ be the set of pair of clusters in $\Pcal^{(i)}$ which are subsets 
of different ground-truth clusters. 
Then, define the \explink loss function as follows:
{\small
\begin{equation} \label{eq:expLoss}
 J(\theta,\alpha) = \sum_{i=1}^{n'} \smashoperator[r]{\sum_{\mathbf{C}_{u,v} \in \Pcal_{-}^{(i)}}} \max\Big\{0 , \expLinkFunc{\alpha}(\mathbf{C}_{u',v'}) -  \expLinkFunc{\alpha}(\mathbf{C}_{u,v}) \Big\}
\end{equation}}
where,
$\mathbf{C}_{u',v'} = \argmin_{\mathbf{C}_{u,v} \in \Pcal_{+}^{(i)} }
\expLinkFunc{\alpha}(\mathbf{C}_{u,v})$ and $n'$ is the round of \HAC
after which no pure merger exists
(i.e., $\forall n > n',$ $\Pcal_{+}^{(n)}=\emptyset$). 
In words, a loss is incurred when two clusters which 
are subset of different ground-truth clusters would be merged 
when a pure merger i.e. a pair of clusters which both belong to 
the same ground-truth cluster also exists. The training procedure
starts with initializing \explink with a particular linkage by
randomly picking a value of $\alpha$. In round $i$ of \HAC,
find $\mathbf{C}_{u',v'} \in \Pcal_{+}^{(i)}$, 
the closest pair of clusters with respect to
$\expLinkFunc{\alpha}$ such that $C_{u'}$ and $C_{v'}$ are subsets of the
\emph{same} ground-truth cluster. 
Then, if a pair of clusters $C_{u,v} \in
\Pcal_{-}^{(i)}$, whose data points belong to \emph{different} ground-truth
clusters, has \emph{smaller} linkage cost than $\mathbf{C}_{u',v'}$, then compute gradients to increase the linkage cost of $C_{u,v}$ and to decrease the linkage cost of $\mathbf{C}_{u',v'}$. 
Then, merge $C_{u'}$ and $C_{v'}$ and repeat this
procedure in round $i+1$ if $\Pcal^{(i+1)} \neq \emptyset$. 
Note that $\mathbf{C}_{u',v'}$ need not be
the least dissimilar pair of clusters to merge as per $\expLinkFunc{\alpha}$ as long as
both $C_{u'}$ and $C_{v'}$ belong to the same ground-truth cluster and we
never perform an \emph{impure merger} during training. \hac rounds
terminate when no pure mergers remain after which parameters are updated w.r.t the
loss in Eq.~\ref{eq:expLoss} in order to make pure mergers more preferable
than competing impure mergers. Pseudocode appears in Algorithm~\ref{alg:genTrainExpLink}.

\setlength{\textfloatsep}{2pt}
\begin{algorithm}[t]
   \caption{\texttt{train\_ExpLink}$(\data, \Ccal^\star, T, \gamma_1,\gamma_2)$}
   \label{alg:genTrainExpLink}
\begin{algorithmic}
	\STATE \textbf{Init}: $\theta$, $\alpha$
	\FOR{$t = 1,\ \dots,\ T$}
		\STATE $J \gets 0$
		\STATE $\Tcal^{(0)}_j \gets \{x_j\} \ \ \ \ \forall \ x_j \in \data$
		\FOR{round $i = 1,\ \dots, n'$} 
			\STATE $\{\Tcal_k^{(i)}\}_k^{l_i} \gets \text{\hac-Round}(\{\Tcal_k^{(i-1)}\}_k^{l_{i-1}})$ 
			\STATE $\{C^{(i)}\}_k^{l_i} \gets \{\lvs{\Tcal_k^{(i)}}\}_k^{l_i} $
			\STATE $\Ccal^{(i)} \gets \{C^{(i)}\}_k^{l_i} $
		
			\STATE $\Pcal^{(i)} \gets  \{\mathbf{C}_{u,v} \in \Ccal^{(i)}\times\Ccal^{(i)} : C_u \neq C_v \} $
			\STATE $\Pcal_{+}^{(i)} \gets \{\mathbf{C}_{u,v} \in \Pcal^{(i)} : \exists C^\star_j \text{ s.t. } C_u,C_v \subset C^\star_j \} $
			\STATE $\Pcal_{-}^{(i)} \gets \Pcal^{(i)} \setminus \Pcal_{+}^{(i)}$
			\STATE $\mathbf{C}_{u',v'} \gets \argmin_{\mathbf{C}_{u,v} \in \Pcal_{+}^{(i)}} 		\expLinkFunc{\alpha}(\mathbf{C}_{u,v})$
			\FOR{$\mathbf{C}_{u,v} \in \Pcal_{-}^{(i)}$ }
				\STATE $J \gets J + \max\Big\{0 , \expLinkFunc{\alpha}(\mathbf{C}_{u',v'}) -  \expLinkFunc{\alpha}(\mathbf{C}_{u,v}) \Big\}$
			\ENDFOR
		\ENDFOR
		\STATE $\theta \gets \theta - \gamma_1 \frac{\partial J}{\partial \theta}$
		\STATE $\alpha \gets \alpha - \gamma_2 \frac{\partial J}{\partial \alpha}$
	\ENDFOR
\end{algorithmic}
\end{algorithm}

Empirically, we find that training is more robust when the loss
function is augmented with a fixed threshold $\tau \in \RR$, and
margin $\mu \in \RR$.  That is, a pure merger incurs a loss only if the
two clusters participating in the merge have dissimilarity greater
than $\tau - \mu$; an impure merger incurs a loss only if the two clusters have dissimilarity less than $\tau +
\mu$.

In order to train the dissimilarity function for a particular member of 
\explink family, gradient updates to $\theta$, parameters of
 the dissimilarity function, are performed to minimize loss in 
 Eq.~\ref{eq:expLoss} while keeping $\alpha$ fixed.
Similarly, the most suitable member of \explink family for
a given dissimilarity function is chosen by training $\alpha$ to minimize loss in 
Eq.~\ref{eq:expLoss} while keeping the dissimilarity function 
parameters fixed.

As anecdotal evidence, we note that the 
linkage function from \explink family used to cluster the data in Figure~\ref{fig:pathExpLink}
was learned using this training procedure while by minimizing the loss on
the same data.

\section{Experiments}
\label{sec:exp}

We experiment with the cross-product of four linkage
functions--single~(\singleLinkTitle), average~(\avgLinkTitle),
complete~(\compLinkTitle) and \explink~(\expLinkTitle)--and the
following eight algorithms for learning pairwise dissimilarity function ($f_\theta$):

\begin{itemize}[noitemsep,topsep=0pt,parsep=0pt,partopsep=0pt,leftmargin=*]
\item \textbf{\AllPairs} (\apExpSec): uses all 
  within- and across-cluster data point pairs to train $f_\theta$.
\item \textbf{\Triplet} (\trp): sample a point $x_i \in
  \data$. Then, sample points $x_i^{+}$ and $x_i^{-}$ to form
  within- and across-cluster pairs respectively. We generate $N=100 \cdot \lvert \data \rvert$ such
  samples.
\item \textbf{\BestEdges} (\bst): for each $x \in \data$,
  generate within- and across cluster pairs by finding the most and
  least similar data points to it respectively, according to $f_{\theta}$.
\item \mst : proposed training method for \slc (\S\ref{subsec:sl-learn}).
\item \expMinus, \expZero, \expPlus proposed training procedure
 for {\explink} with fixed $\alpha$ (\S\ref{subsec:expfun_train}). \expMinus refers to $\alpha= -\infty$, \expZero to  $\alpha=0$, and \expPlus to $\alpha=\infty$.
\item \expAlpha: proposed training procedure for \explink
with joint learning of $\alpha$ and $f_\theta$ (\S \ref{subsec:expfun_train}).
\end{itemize}

Our experiment is designed to determine if the best clustering algorithm depends on the dataset or if 
there is a universally top performing algorithm.
Additionally, our experiment tests whether 
the performance of a linkage function depends on the training algorithm.
We randomly divide each dataset into 50
train/dev/test splits. For each linkage and training algorithm pair,
for each split, we learn a dissimilarity function on the training set,
tune on the development set and cluster the test set. We record the
performance on each split. To compare two training algorithms for the
same linkage, we compute their mean performance over all splits.
When using \expLinkTitle as the clustering algorithm with the dissimilarity function trained using a training method other than \expAlpha, the most appropriate linkage from \explink is chosen \emph{after} learning the dissimilarity function by minimizing loss in Equation~\ref{eq:expLoss} on training data with respect to $\alpha$.

We conduct experiments with the following four datasets:

\begin{itemize}[noitemsep,topsep=0pt,parsep=0pt,partopsep=0pt,leftmargin=*]
\item \textbf{UMIST Face Data (Faces)}~\cite{graham1998characterising} :  
  564 gray-scale images with 20 ground-truth clusters. The
  pre-cropped images are downsampled to $56 \times 46$. We use PCA to
  reduce the data to 20 dimensions. 7 clusters are used for training, 6
  for dev, and 7 for test set.
\item \textbf{Noun Phrase Coreference (NP
    Coref)}~\cite{hasler2006nps}: 104 documents, each contains
  clusters of coreferent noun phrases (NPs). Each pair of NPs is a
  described by 102-dim vector~\cite{stoyanov2009conundrums}. We use 62
  documents for training, 10 for dev, and 32 for test.
\item \textbf{Rexa}~\cite{culotta2007author}: 1459 bibliographic
  records of authors divided into 8 blocks w.r.t. unique first initial and last
  name. Each pair of records within a block is represented using 14-dim vector.  
  We use 3 blocks for training, 2 for dev and 3 for test.
\item \textbf{AMINER}~\cite{wang2011adana}: 6730 publication records
  of authors divided into 100 blocks. Each pair of publications within
  a block is represented by a 8-dim vector~\cite{wang2011adana}.
  We use 60 blocks for training, 10 for dev, and 30 for test.
\end{itemize}

For Faces, we learn Mahalanobis distance matrix $M \succeq 0$.
To do so, we learn a matrix $A$ s.t. $M = A^{T}A$. We use threshold, $\tau=100$, and
margin, $\mu=10$, when computing the loss. For other datasets, we use
feature vectors representing data point pairs and train an average
perceptron for the dissimilarity function with threshold, $\tau=0$,
and margin, $\mu=2$. Code for experiments is available at: \url{https://github.com/iesl/expLinkage}.
\subsection{Hierarchical Clustering Evaluation}

\paragraph{Dendrogram Purity} The output of hierarchical clustering 
algorithms is a cluster tree, rather than a flat clustering. 
Following approach used in~\citet{heller2005bayesian}, we evaluate 
cluster trees using dendrogram purity, which is a holistic 
measure of the tree quality. Given $\Tcal$, a hierarchical 
clustering of $\data=\{x_i\}_{i=1}^m$,
with ground-truth clusters $\Ccal^\star$, the dendrogram purity of
$\Tcal$ is:
\begin{align*}
\texttt{DP}(\Tcal) = \frac{1}{|\Wcal^\star|}\sum_{x_i,x_j \in  \Wcal^\star } \texttt{pur}(\lvs{\texttt{LCA}(x_i,x_j)}, \mathcal{C}^\star(x_i))
\end{align*}
where $\mathcal{C}^\star(x_i)$ 
gives the ground-truth cluster of the point~$x_i$,
 $\Wcal^\star$ is the set of unordered pairs of points 
belonging to the same ground-truth cluster,
$\texttt{LCA}(x_i,x_j)$ is the lowest common ancestor of $x_i$
and $x_j$ in $\mathcal{T}$, $\lvs{z}\subset \data$ is the set of
leaves for any internal node $z$ in $\mathcal{T}$, and
\texttt{pur}$(S_1,S_2) = |S_1 \cap S_2|/|S_1|$.

In words, to compute dendrogram purity of a tree $\Tcal$ with respect to
ground-truth clusters $\Ccal^\star$,  iterate
over all pairs of points $(x_i,x_j)$ which belong to the same ground-truth
cluster, find the smallest subtree containing $x_i$ and $x_j$, and measure
fraction of leaves in that subtree which are in the same ground-truth
cluster as $x_i$ and $x_j$.

\begin{figure*}[t]
	\begin{centering}
		\captionsetup[subfigure]{justification=centering}
		\begin{subfigure}[b]{1.0\textwidth}
\centering
{\scriptsize
\begin{tabular}{cccccccccccccccccccc}
\toprule
\multirow{2}{*}{Obj} & \multicolumn{4}{c}{Rexa} & &\multicolumn{4}{c}{AMINER} & & \multicolumn{4}{c}{NP Coref}  & & \multicolumn{4}{c}{Faces} \\\cline{2-5} \cline{7-10} \cline{12-15} \cline{17-20}
 			& 	\singleLinkTitle &  \avgLinkTitle 	& 	\compLinkTitle & 	\expLinkTitle 	& & 	\singleLinkTitle &  \avgLinkTitle 	& 	\compLinkTitle & 	\expLinkTitle 	 & & 	\singleLinkTitle &  \avgLinkTitle 	& 	\compLinkTitle & 	\expLinkTitle 	 & & 	\singleLinkTitle &  \avgLinkTitle 	& 	\compLinkTitle & 	\expLinkTitle 	\\
\midrule
\bst 		& 86.9                                     & \underline{\underline{84.0}}             & \underline{\underline{75.5}}             & \underline{\underline{87.8}}            & & \underline{\underline{93.2}}             & \underline{\underline{93.6}}             & \textbf{82.0}                            & \underline{\underline{93.5}}            & & \underline{\underline{60.5}}             & \underline{\underline{57.5}}             & \underline{\underline{46.8}}             & \underline{\underline{52.8}}            & & \underline{\underline{93.7}}             & \underline{\underline{77.4}}             & \underline{\underline{70.9}}             & \underline{\underline{93.7}}            \\
\mst 		& \textbf{87.3}                            & \underline{\underline{85.1}}             & \underline{\underline{75.5}}             & \underline{\underline{88.4}}            & & \underline{\underline{92.7}}             & \underline{\underline{93.1}}             & \underline{\underline{81.5}}             & \underline{\underline{93.2}}            & & \underline{\underline{59.1}}             & \underline{\underline{54.9}}             & \underline{\underline{47.6}}             & \underline{\underline{53.6}}            & & \textbf{95.3}                            & \underline{\underline{81.7}}             & \underline{\underline{76.2}}             & \textbf{95.4}                           \\
\expMinus	& \textbf{87.3}                            & \underline{\underline{85.6}}             & \underline{\underline{76.2}}             & \underline{88.6}                        & & \underline{\underline{87.6}}             & \underline{\underline{84.9}}             & \underline{\underline{77.8}}             & \underline{\underline{85.3}}            & & \textbf{63.3}                            & \underline{\underline{62.0}}             & \underline{\underline{55.8}}             & \textbf{64.3}                           & & \underline{94.5}                         & \underline{\underline{79.9}}             & \underline{\underline{74.9}}             & \underline{94.6}                        \\
\apExpSec  		& \underline{\underline{81.7}}             & \underline{\underline{84.6}}             & \underline{\underline{80.1}}             & \underline{\underline{82.3}}            & & \underline{\underline{92.8}}             & \underline{\underline{93.4}}             & \underline{\underline{81.8}}             & \underline{\underline{93.4}}            & & \underline{\underline{58.7}}             & \underline{\underline{58.1}}             & \underline{\underline{50.9}}             & \underline{\underline{55.3}}            & & \underline{\underline{91.3}}             & 85.6                                     & 81.7                                     & \underline{\underline{86.3}}            \\
\trp 		& \underline{\underline{85.6}}             & \underline{88.1}                         & \textbf{82.4}                            & 89.1                                    & & \underline{\underline{92.0}}             & \underline{\underline{93.2}}             & \underline{\underline{81.1}}             & \underline{\underline{92.9}}            & & \underline{\underline{59.3}}             & \underline{\underline{61.6}}             & \underline{\underline{56.6}}             & \underline{\underline{62.2}}            & & \underline{\underline{91.0}}             & \textbf{85.8}                            & \textbf{82.2}                            & \underline{\underline{85.8}}            \\
\expZero	& \underline{\underline{85.5}}             & \textbf{88.9}                            & \underline{\underline{79.1}}             & \textbf{89.5}                           & & \textbf{93.4}                            & \textbf{93.9}                            & \underline{\underline{81.7}}             & \textbf{94.1}                           & & \underline{\underline{61.0}}             & \textbf{62.8}                            & \textbf{57.4}                            & \underline{\underline{63.5}}            & & \underline{\underline{91.0}}             & 84.8                                     & 80.6                                     & \underline{\underline{90.6}}            \\
\expPlus 	& \underline{\underline{83.9}}             & \underline{\underline{87.3}}             & 81.8                                     & \underline{\underline{88.1}}            & & \underline{\underline{92.4}}             & \underline{\underline{92.7}}             & \underline{\underline{81.0}}             & \underline{\underline{90.7}}            & & \underline{\underline{61.9}}             & \textbf{62.8}                            & \underline{\underline{56.6}}             & \underline{\underline{62.5}}            & & \underline{\underline{90.4}}             & \underline{84.1}                         & 80.5                                     & \underline{\underline{83.4}}            \\
\expAlpha 	& 87.1                                     & \underline{\underline{86.9}}             & \underline{\underline{76.4}}             & 89.1                                    & & \textbf{93.4}                            & \textbf{93.9}                            & \underline{\underline{81.7}}             & \textbf{94.1}                           & & \underline{\underline{61.2}}             & \textbf{62.8}                            & 57.3                                     & \underline{\underline{63.4}}            & & \underline{\underline{94.2}}             & \underline{\underline{79.2}}             & \underline{\underline{73.9}}             & \underline{94.5}                        \\

\bottomrule
\end{tabular}
}	
\caption{Dendrogram Purity for all training methods}
\label{table:dendPurity}
\end{subfigure} 
		\begin{subfigure}[b]{1.0\textwidth}
	\centering
{	\scriptsize
	\begin{tabular}{ccccccccccccccccccccccc}
		\toprule
		\multirow{2}{*}{Obj} & 				\multicolumn{4}{c}{Rexa} 								& &					\multicolumn{4}{c}{AMINER} 										& & 						\multicolumn{4}{c}{NP Coref}  							& & 				\multicolumn{4}{c}{Faces} 		\\\cline{2-5} \cline{7-10} \cline{12-15} \cline{17-20}
		& 	\singleLinkTitle &  \avgLinkTitle 	& 	\compLinkTitle & 	\expLinkTitle 	& & 	\singleLinkTitle &  \avgLinkTitle 	& 	\compLinkTitle & 	\expLinkTitle 	 & & 	\singleLinkTitle &  \avgLinkTitle 	& 	\compLinkTitle & 	\expLinkTitle 	 & & 	\singleLinkTitle &  \avgLinkTitle 	& 	\compLinkTitle & 	\expLinkTitle 	\\
		\midrule

\bst      	& 75.0                                     & \underline{\underline{55.4}}             & \underline{\underline{38.9}}             & \underline{74.2}                        &  & \textbf{79.3}                            & \textbf{78.7}                            & \textbf{43.2}                            & \textbf{81.0}                           &  & 50.0                                     & \underline{\underline{44.3}}             & \underline{\underline{37.1}}             & \underline{\underline{39.9}}            &  & \textbf{77.6}                            & \underline{\underline{60.2}}             & \underline{\underline{54.6}}             & \textbf{78.8}                           \\
\mst      	& \textbf{75.9}                            & \underline{\underline{56.9}}             & \underline{\underline{41.0}}             & \underline{73.4}                        &  & 79.1                                     & 77.9                                     & 42.1                                     & \textbf{81.0}                           &  & \underline{\underline{48.2}}             & \underline{\underline{43.3}}             & \underline{\underline{37.5}}             & \underline{\underline{42.6}}            &  & 76.9                                     & \underline{\underline{63.9}}             & \underline{\underline{59.2}}             & 77.7                                    \\
\expMinus 	& 74.0                                     & \underline{\underline{60.8}}             & \underline{\underline{43.4}}             & 76.7                                    &  & \underline{\underline{75.1}}             & \underline{\underline{66.9}}             & 43.1                                     & \underline{\underline{68.3}}            &  & 50.2                                     & \underline{\underline{44.5}}             & \underline{\underline{38.0}}             & 47.6                                    &  & 73.1                                     & \underline{\underline{60.8}}             & \underline{\underline{58.9}}             & 74.5                                    \\
\apExpSec        	& \underline{\underline{59.8}}             & \underline{\underline{58.0}}             & \underline{\underline{49.4}}             & \underline{\underline{56.9}}            &  & \underline{\underline{75.9}}             & 76.4                                     & 42.9                                     & \underline{\underline{75.3}}            &  & \underline{\underline{46.4}}             & \underline{46.7}                         & 41.1                                     & \underline{\underline{43.9}}            &  & \underline{\underline{70.2}}             & 68.6                                     & 65.9                                     & \underline{\underline{69.4}}            \\
\trp  		& \underline{\underline{66.9}}             & \underline{\underline{61.2}}             & \textbf{56.5}                            & \underline{\underline{71.7}}            &  & \underline{\underline{73.9}}             & \underline{75.3}                         & 41.8                                     & \underline{\underline{77.2}}            &  & \underline{\underline{39.7}}             & \underline{\underline{45.9}}             & \underline{\underline{38.9}}             & \underline{\underline{46.0}}            &  & \underline{\underline{70.6}}             & \textbf{68.8}                            & \textbf{66.2}                            & \underline{\underline{69.4}}            \\
\expZero 	& \underline{\underline{70.5}}             & \textbf{69.8}                            & \underline{\underline{47.0}}             & \underline{\underline{72.3}}            &  & \underline{76.3}                         & \underline{\underline{69.8}}             & 42.6                                     & \underline{\underline{74.8}}            &  & \underline{\underline{45.8}}             & \textbf{47.8}                            & \textbf{41.9}                            & \textbf{48.1}                           &  & \underline{\underline{69.1}}             & 68.1                                     & 64.6                                     & 74.5                                    \\
\expPlus 	& \underline{\underline{67.5}}             & \underline{\underline{65.0}}             & \underline{\underline{50.0}}             & \underline{\underline{69.5}}            &  & \underline{\underline{74.2}}             & \underline{\underline{69.3}}             & \textbf{43.2}                            & \underline{\underline{59.7}}            &  & \textbf{50.8}                            & 47.7                                     & 41.2                                     & \underline{47.3}                        &  & \underline{\underline{67.7}}             & 66.4                                     & 63.9                                     & \underline{\underline{68.0}}            \\
\expAlpha 	& 74.7                                     & \underline{\underline{63.5}}             & \underline{\underline{42.8}}             & \textbf{78.1}                           &  & \underline{\underline{75.7}}             & \underline{\underline{69.5}}             & 42.6                                     & \underline{\underline{75.4}}            &  & \underline{\underline{46.0}}             & 47.6                                     & 41.5                                     & 47.9                                    &  & 75.0                                     & \underline{\underline{62.1}}             & \underline{\underline{57.6}}             & 75.5                                    \\

		\bottomrule
	\end{tabular}
}
\caption{Pairwise F1 Score for all training methods}
\label{table:pairwiseF1}
\end{subfigure}
 
		\vspace{-7mm}
		\caption{Performance of each training method-linkage pair. Each row
			corresponds to a training algorithm and each column corresponds to a
			linkage function and dataset. Each value represents the mean
			performance of a training algorithm for a linkage
			over 50 train/dev/test splits. Bold numbers indicate the best performing
			training method for a particular linkage. Single underline indicates that
			the value is statistically significantly worse than the best performing method
			with $p<0.05$ and double underline indicate $p<0.01$.}
		\label{table:combinedTable}
	\end{centering}
\end{figure*}

Table~\ref{table:dendPurity} shows mean dendrogram purity of hierarchical clustering for
each training method/linkage function pair on four datasets, averaged over the 50 randomly generated train/dev/test splits. 
In the table, each row represents a training algorithm, and each column a
linkage function and dataset. 
Bold values in each
column indicates the best training method for linkage function and dataset
corresponding to the column. Values with a single underline indicates that the difference 
in performance w.r.t the best training method in the column is statistically 
significant with p-value~$< 0.05$. A double underline is used to
indicate statistical significance with p-value~$< 0.01$, where statistical
significance is measured using resampled paired-t test~\cite{dietterich1998approximate}.

\paragraph{Top Performers.}
For \singleLinkTitle, we find that training with a corresponding method like
\mst, \bst or \expMinus is best on three out of four
datasets\footnote{ \mst and \bst have minor
  differences: every positive example generated by \bst is
  also generated by \mst, but \mst may also include examples to
  guarantee that the positive training examples are the edges in an
  minimum spanning tree over the data.} \footnote{\expMinus differs from \mst in that
  \mst considers only the best across cluster edge for every point,
  while \expMinus might consider more than one across cluster for
  every point.}. 
For \avgLinkTitle, training with \expZero is best
except on the Faces dataset, where training with \trp does marginally better
than training with \expZero.
For \textsc{comp}, we find no recurring
pattern except that it achieves much lower dendrogram purity than any
of the other linkages.  Finally, for \expLinkTitle, joint training with \expAlpha
outperforms other training methods and standard linkages on AMINER, NP Coref and joint training with \expAlpha is only marginally outperformed on Rexa and Faces when choosing a linkage function from \explink \emph{after} training the dissimilarity function. 
Overall, these results suggest matching the training algorithm and 
linkage function to achieve the best performance.

\paragraph{All Pairs.}
Table~\ref{table:dendPurity} also reveals that the training method most commonly used in
practice, \apExpSec, is rarely a good choice. For \singleLinkTitle and \expLinkTitle, \apExpSec
is always worse than the top performer by a statistically significant
margin--on Rexa, NP Coref and Faces, \apExpSec causes a ~4-8\%
drop in dendrogram purity for \singleLinkTitle and \expLinkTitle. 
Despite \avgLinkTitle being the closest match for  \apExpSec among the standard
linkages, \apExpSec is always worse than the top performer for \avgLinkTitle. 
\vspace{-.30cm}
\paragraph{Joint Training.}
A particularly notable result is that jointly learning a linkage from
the \explink family and a corresponding dissimilarity function is best
or closely tracks the best performer on all datasets. Again, on Faces,
training with a method that matches with \slc gives best results.
Joint training does not
require the practitioner to choose a linkage function \emph{a priori}, which is the 
desired setting for real-world data. Our results suggest that joint training is
at least as effective as choosing the best of the standard linkage function/training algorithm pairs. Additionally, joint training is as effective or competitive with choosing a linkage from the \explink family \emph{after} training the dissimilarity function. 

\subsection{Flat Clustering Evaluation}
\paragraph{Pairwise F1.} We evaluate flat clusterings using
pairwise F-Measure (F1)~\cite{manning2010introduction}. Let 
$\Wcal^\star$ be pairs of points that
belong to the same ground-truth cluster, and $\hat{\Wcal}$ be pairs of
points in that belong the same predicted cluster. 
 A true positive is
defined as a pair of points that belong to both $\Wcal^\star$ and
$\hat{\Wcal}$, a true negative belongs to $\Wcal^\star$ but not to
$\hat{\Wcal}$. Similarly, define false positive, and false
negatives. Then, compute pairwise F1 as the harmonic mean of
pairwise precision and recall.
\paragraph{Selecting a flat clustering with a threshold.} 
To extract a flat clustering from a tree $\Tcal$ given a 
threshold value $\xi$, we select the clusters represented by the roots 
of subtrees for which all linkages between siblings in 
the subtree are less than $\xi$ and the linkage of the 
subtree root with its sibling is greater than $\xi$. 
We select a threshold value that leads to the clustering with maximum 
pairwise F1 score on dev set.
Note that choosing a single partition adds another opportunity to
introduce error for all methods (i.e. if the threshold returns a poor
partition from a tree with high dendrogram purity).

Table~\ref{table:pairwiseF1} shows mean pairwise-F1 scores 
for flat clusterings extracted from the trees for 50 random train/dev/test splits
(similar to Table~\ref{table:dendPurity}). Some trends exists
that are similar to those observed with respect to dendrogram purity,
albeit with more exceptions. Generally, a linkage function achieves
best performance when dissimilarity is learned using a matching
training algorithm.  \singleLinkTitle is dominant on Faces, as is the case with
dendrogram purity.  The magnitude of relative differences between the
methods are larger, for example: on Rexa, when clustering with
\expLinkTitle, joint training leads to more than 20 points higher F1 than \apExpSec
training, and on AMINER, best performing method is better than
\expPlus by ~20 points F1. While these results are informative, they depend largely on
the method for selecting the tree consistent partition. Methods that
select better tree-consistent partitions (than the threshold method) are likely to exist.
\section{Related Work}

Several approaches have been proposed to learn a distance metric to
optimize performance of classification methods such as k-NN and
clustering methods such as k-means
\cite{xing2003distance,goldberger2005neighbourhood,globerson2006metric,
kunapuli2012mirror,weinberger2009distance,ashtiani2015representation}. 
These approaches operate under the semi-supervised setting where the
metric is learned using a small fraction of within- and across-cluster
pairs, or using clustering of a small fraction of data points 
and the learned distance metric is used on the
\emph{same} set of clusters. 
\citet{balcan2008clustering,awasthi2010supervised} operate in the setting 
where the goal is to learn the desired clustering with help of an oracle
with the goal of minimizing total number of queries to the oracle. 
Hierarchical clustering algorithms have
also received much attention in trying to incorporate
constraints/labels on a small fraction of
points \cite{zheng2011semi,li2011hierarchical,xiao2016semi, 
chatziafratis2018hierarchical}.  Our work differs from these in
that we operate in the supervised setting where we
use \emph{all} labels in the training dataset to learn a pairwise dissimilarity
and linkage function for \hac, that
generalizes to \emph{different} test set of clusters. 

For the task of noun-phrase coreference, several heuristic 
approaches for generating training examples have been
proposed~\cite{soon2001machine,ng2002combining,ng2002improving}. Recent
approaches for coreference resolution use latent tree models, where
each noun-phrase is linked to its closest/best proceeding
noun-phrase~\cite{lassalle2015joint,chang2013constrained,durrett2013easy}. Our
training procedure for \slc differs from these approaches
in that we do not make use of the word order when constructing an MST
over ground-truth clusters, and since these approaches rely on the
word order in raw text, they are not applicable to other supervised
clustering tasks such as author coreference. The most related to our training procedure for SL is \citet{yu2009learning}, in
which the authors learn a latent tree model using SVMs for noun-phrase
coreference. Whereas their loss function
penalizes the merger of two different
ground-truth clusters, ours also penalizes of splitting ground-truth
clusters.  Unlike our work, they only use \slc while our training
procedure optimizes over the \explink family.

\citet{culotta2007author} present an approach for generating training
examples using errors produced during \hac. \citet{culotta2007author}
use a loss that only considers the impure agglomeration 
that appears in the earliest round of \hac whereas our work uses
a loss function based on multiple rounds of a modified version of \hac that 
considers pure agglomerations. Also, \citet{culotta2007author} learn a function that scores sets
and while we learn pairwise dissimilarities.

There is also related work on \emph{unsupervised} hierarchical 
clustering objectives which describes the quality of a 
tree structured clustering of the data and, unlike the supervised 
objectives described in this paper, assume that a
similarity or dissimilarity function is given~\cite{dasgupta2016cost,roy2016hierarchical,
cohen2017hierarchical,moseley2017approximation,charikar2017approximate,
cohen2018hierarchical,charikar2019hierarchical}.
 
\section{Conclusion}
In this paper, we examine the dependence between training methods and
\hac variants in the supervised clustering setting. Using the popular \hac with
\slc algorithm as an example, we show that mismatch between
training and clustering objectives leads to poor performance. Then, we
present a training algorithm suited specifically for \hac with \slc
that yields improved results. We introduce a new family of \hac
linkage functions that smoothly interpolates between single, average
and complete linkage---called the Exponential Linkage (\explink)
family---and provide a joint training algorithm that simultaneously
learns an appropriate linkage in the \explink family and a corresponding
pairwise dissimilarity function.  In experiments, we demonstrate that
\explink, coupled with our training algorithm, outperforms or is 
competitive with the best \hac variant--an important
result for practitioners since the best \hac variant for a problem at
hand is often unknown. Our experiments also underscore the notion that
matching training and clustering objectives leads to superior test
time performance in the supervised clustering setting.

\subsection*{Acknowledgments} 
We thank Ian Gemp and Javier Burroni for many helpful discussions and feedback. We also thank the anonymous reviewers for their constructive feedback. 
This work was supported in part by the Center for Data Science and the Center for Intelligent Information Retrieval, in part by the Chan Zuckerberg Initiative under the project “Scientific Knowledge Base Construction, and in part by the National Science Foundation under Grant No. NSF-1763618. The work reported here was performed in part using high performance computing equipment obtained under a grant from the Collaborative R\&D Fund managed by the Massachusetts Technology Collaborative. Any opinions, findings and conclusions or recommendations expressed in this material are those of the authors and do not necessarily reflect those of the sponsor.

\bibliography{icml_references}
\bibliographystyle{icml2019}

\clearpage
\appendix
\section{Appendix}

\subsection{Analysis of Loss Function for \explink}
\textbf{Fact 1} \emph{Given an injective linkage function $\Psi^\alpha$, performing greedy \hac with \explink-$\alpha$ and dissimilarity 
function $f_\theta$ results in cluster tree with perfect dendrogram purity if the loss, $J(\theta,\alpha)$, given by equation~\ref{eq:expLoss} is zero}
\label{proof:zero_loss_perfect_dp}
\begin{proof}
	Let $\data = \{x_i\}_{i=1}^{m}$ with ground-truth clusters $\Ccal^\star = \{C_i^\star\}_{i=1}^{K}$. Let$\Tcal$ be tree build by \hac on $\data$ using linkage function $\expLinkFunc{\alpha}$ and dissimilarity function $f_\theta$. 

	To prove $J(\theta,\alpha)=0 \implies \texttt{DP}(\Tcal) = 1$, we will prove the contrapositive i.e. \texttt{DP}$(\Tcal) < 1 \implies J(\theta,\alpha) > 0$.

	\texttt{DP}($\Tcal$) is purity of the lowest common ancestor of a pair of points in $\Tcal$, averaged over every pair of points in the same ground-truth cluster.
	If \texttt{DP}$(\Tcal) < 1$, then $\exists C^\star \in \Ccal^\star, x_a,x_b \in C^\star $ s.t. $\texttt{purity}(\texttt{LCA}(x_a,x_b),C^\star)<1$ i.e. $\texttt{LCA}(x_a,x_b)$ is the root node of an impure subtree in $\Tcal$.

	Let $v_{a,b} = \texttt{LCA}(x_a,x_b)$, and let $C_{v_{a,b}}$ be cluster comprised of points at leaves of tree rooted at $v_{a,b}$.
	Since $\texttt{purity}(v_{a,b}, C^\star) < 1$, at least one of $v_{a,b}$'s subtrees is impure. WLOG, suppose the subtree containing $x_a$ is impure. So, there is a descendant $v'$ of $v_{a,b}$ with children $v'_l,v'_r$ such that $x_a \in C_{v'_l}, C_{v'_l} \subset C^\star $, and $C_{v'_r} \not\subset C^\star$. This means that $v'$ is the first impure ancestor of $x_a$.
	
	Let $j$ be the smallest round in which such an impure ancestor of any two points in the same ground-truth cluster is created. Let $x_a$ and $x_b$ be these two points.
	Before round $j$, either every cluster is pure cluster (i.e., a subset of a ground-truth cluster), or an impure cluster formed by the union of several ground-truth clusterss. If there exists an impure cluster in round $j$ other than those formed by the union of several ground-truth clusters, then it contradicts $j$ being the smallest round in which an impure ancestor of any two points in the same ground-truth cluster is created.

	In round $j$, let the impure merge occur between a pure cluster $C_a$ and a cluster $C_b$ where $ \exists C^\star \in \Ccal^\star, C_a \subset C^\star,  C_b \not\subset C^\star$. Since $C_a$ is a strict subset of $C^\star$, there exists at least one more cluster in round $j$ which a strict subset of $C^\star$, and hence there exists at least one pure merger in round $j$.
	Let $C_{a_+,b_+}$ be the best pure merger available in round $j$.

	Since \hac chooses to merge $C_{a,b}$ in round $j$ over $C_{a_+,b_+}$,
	\[ \expLinkFunc{\alpha}(C_{a,b}) \leq \expLinkFunc{\alpha}(C_{a_+,b_+})  \]
	Further, since $\expLinkFunc{\alpha}$ is injective, we have a strict inequality
	\begin{align*}
	\expLinkFunc{\alpha}(C_{a,b}) &< \expLinkFunc{\alpha}(C_{a_+,b_+}) \\
	\implies \expLinkFunc{\alpha}(C_{a_+,b_+}) &- \expLinkFunc{\alpha}(C_{a,b}) > 0
	\end{align*}
	Thus, $J(\theta,\alpha) \geq \max\{0, \expLinkFunc{\alpha}(C_{a_+,b_+}) - \expLinkFunc{\alpha}(C_{a,b}) \} > 0$

	Loss incurred in round $j$ is greater zero because the pure merger available in round $j$ is worse than best impure merger available in round $j$.

\end{proof}

\subsection{Comparison to other inference methods}

Top-down tree construction methods have been shown to be effective
at optimizing unsupervised hierarchical clustering objectives \cite{dasgupta2016cost}.
While there is no natural extension of our training objective for these inference
methods, we provide an empirical comparison between \hac inference and the recursive 
sparsest cut (\RSC) approach with the dissimilarity function trained using different training procedures. 
We implement \RSC \ using scikit-learn's spectral clustering \cite{pedregosa2011scikit}. 

Figure~\ref{table:rsc} shows mean dendrogram purity results for 50 train/test/dev splits. Each row corresponds to a training procedure 
for learning the dissimilarity function. The \hac column contains the best dendrogram purity for hierarchical clustering using \singleLinkTitle, \avgLinkTitle, \compLinkTitle or \expLinkTitle linkage, and the \RSC \ column contains dendrogram purity for top-down hierarchical clustering obtained using recursive sparsest cut.  
The results of this experiments show that the approaches that use an inference procedure
aligned with the training procedure (namely the \hac-based approach presented in this paper)
are always more performant than \RSC. 
\begin{figure}[H]
\centering
 \begin{tabular}{c c c c c c c c }
 	\toprule
 	\multirow{2}{*}{Obj} & \multicolumn{2}{c}{Rexa} & & \multicolumn{2}{c}{AMINER} \\\cline{2-3} \cline{5-6} 
 	& \hac	& \RSC 	& &	\hac 	& \RSC  \\
 	\midrule
 	\bst			& 87.8	& 74.3	& & 93.6 	& 88.8 	\\
 	\mst			& 88.4	& 74.8	& & 93.2	& 88.1 	\\
 	\expMinus		& 88.6	& 73.1	& & 85.3 	& 79.3 	\\
 	\apExpSec		& 84.6	& 75.0	& & 93.4	& 87.9 	\\
 	\trp			& 89.1	& 77.2	& & 93.2	& 87.3 	\\
 	\expZero		& 89.5	& 76.6	& & 94.1	& 81.6 	\\
 	\expPlus		& 88.1	& 76.3	& & 92.7 	& 81.5 	\\
 	\expAlpha		& 89.1	& 75.1	& & 94.1	& 81.5 	\\
 	\bottomrule
 \end{tabular}

 	\begin{tabular}{c c c c c c c c }
	\toprule
	\multirow{2}{*}{Obj} & \multicolumn{2}{c}{NP Coref} & & \multicolumn{2}{c}{Faces} \\\cline{2-3} \cline{5-6} 
	& \hac	& \RSC 	& &	\hac 	& \RSC  \\
	\midrule
	\bst			& 	60.5	& 32.9	& & 93.7	& 69.6	\\
	\mst			& 	59.1	& 37.6	& & 95.4	& 74.7	\\
	\expMinus		& 	64.3	& 49.3	& & 94.6	& 73.6	\\
	\apExpSec		& 	58.7	& 39.7	& & 91.3	& 81.0	\\
	\trp			& 	62.2	& 54.1	& & 91.0	& 81.0	\\
	\expZero		& 	63.5	& 50.5	& & 91.0	& 78.5	\\
	\expPlus		& 	62.8	& 52.6	& & 90.4	& 78.7	\\
	\expAlpha		& 	63.4	& 50.4	& & 94.5	& 72.9	\\
	\bottomrule
\end{tabular}
 \caption{Dendrogram purity results for \RSC \ and \hac with best linkage function for eight training methods.}
 \label{table:rsc}
\end{figure}





\end{document}